\documentclass[conference]{IEEEtran}
\IEEEoverridecommandlockouts
\usepackage{cite}
\usepackage{amsmath,amssymb,amsfonts}
\usepackage{algorithmic}
\usepackage{graphicx}
\usepackage{textcomp}
\usepackage{array}
\usepackage{xcolor}
\usepackage{listings}
\usepackage{float}
\usepackage{comment}
\usepackage{pgfplots}
\def\BibTeX{{\rm B\kern-.05em{\sc i\kern-.025em b}\kern-.08em
    T\kern-.1667em\lower.7ex\hbox{E}\kern-.125emX}}

\usepackage{subcaption}   

\begin{document}

\bibliographystyle{IEEEtran}

\title{Intentional Biases in LLM Responses\\

}

\author{\IEEEauthorblockN{1\textsuperscript{st} Nicklaus Badyal}
\IEEEauthorblockA{\textit{dept. of Computer Science} \\
\textit{Univ. of Victoria}\\
Victoria, Canada\\
nicklausb@uvic.ca}
\and
\IEEEauthorblockN{2\textsuperscript{nd} Derek Jacoby}
\IEEEauthorblockA{\textit{dept. of Computer Science} \\
\textit{Univ. of Victoria}\\
Victoria, Canada \\
0000-0002-1552-7484}
\and
\IEEEauthorblockN{3\textsuperscript{rd} Yvonne Coady}
\IEEEauthorblockA{\textit{dept. of Computer Science} \\
\textit{Univ. of Victoria}\\
Victoria, Canada \\
ycoady@uvic.ca}
}

\maketitle

\begin{abstract}

In this study we intentionally introduce biases into large language model responses in an attempt to create specific personas for interactive media purposes. We explore the differences between open source models such as Falcon-7b and the GPT-4 model from Open AI, and we quantify some differences in responses afforded by the two systems. We find that the guardrails in the GPT-4 mixture of experts models with a supervisor, while useful in assuring AI alignment in general, are detrimental in trying to construct personas with a variety of uncommon viewpoints. This study aims to set the groundwork for future exploration in intentional biases of large language models such that these practices can be applied in the creative field, and new forms of media.

\end{abstract}


\section{Introduction}

The rapid growth of natural language processing and large language models has created numerous opportunities for transformative human computer interaction experiences that offer interactive, contextual, and example based explanations. These types of experiences promise to greatly boost learning for students, create new mediums for educators, and allow for creative interpretations and reenactments of personas to be performed by large language models (LLMs). However, as found in \cite{kolisko_exploring_2023} and \cite{penedo_refinedweb_2023} models can contain bias in the way they respond, as well as contain traces of toxicity in their datasets. Our goal is to create a climate AI avatar replicating what someone like David Suzuki or Greta Thunberg might say to explain climate change. This led us into the question of how can you bias a language model to form a specific persona, and how does this capability vary across types of LLM models.

We initially used contextual interviews with survivors of climate disasters as our source materials, but found that the materials did not provide enough breadth to form a general persona. In choosing climate textbooks to supplement these interviews we decided to see how far we could bias a language model outside the constraints of its general training materials. We found that with the proper prompting and source material a Generative Pretrained Transformer (GPT) model could be heavily biased into conveying false information, and hallucinating incorrect examples that were not in the training data to make it's points. Using The Heartland Institutes' "Climate at a Glance textbook" \cite{noauthor_heartland_nodate}, which contains heavily biased information on climate change and it's effects, we were able to consistently reproduce misinformation on climate change in relation to wildfires, flooding, icecaps melting, etc. On the way to making Greta-in-a-box, we decided to make an anti-Greta to explore questions of bias and focus in our AI avatar. This proved more challenging since our anti-Greta persona context was in conflict with much of the background training material for the LLM.

Due to the OpenAI GPT models having significant guardrails, we hypothesized that an open source model without guardrails would be more easily biased towards a less common viewpoint when given prompts designed to do so. For this test, we constructed context and question pairs that were either in alignment, out of alignment, or neutral with respect to the background materials the model was trained upon. We further classified these questions as high coverage, meaning that the background training material could be reasonably expected to contain a correct answer, or low coverage in which the answer to the question would exist only in the provided context. We then asked these questions to GPT-4, and the open source Falcon-7b instruct model. The guardrails contained in a mixture-of-experts model like GPT make it less likely to generate an answer that is out of concordance with its background training material, which reduces its likelihood of hallucinating, but also restricts the range of personalities that the model can realistically represent. We show some of these constraints in our attempts to bias models in ways unsupported by the background training materials.

In the following sections we will outline some related work and then present our results of attempting to intentionally bias the GPT and Falcon LLMs before concluding with some discussion of situations in which biases are essential for creative purposes.

\section{Related work}

In this section, we examine related work on LLM model hallucinations, on mixture-of-experts models like Chat-GPT versus dense models like Falcon, and examine prompting strategies and other means of creating a consistent persona in LLM responses.

\subsection{Model architectures}

Although OpenAI has not released a definitive architectural paper on the GPT family of models, there is some evidence that it is structured as a mixture of experts model which consists of a number of smaller models (rumored to be 8, in this case) with a supervisor to choose the best result \cite{romero_gpt-4s_2023}. This is a structure with some definite advantages, both computationally, and in terms of being able to construct guardrails to try to ensure well-aligned responses from the model \cite{artetxe_efficient_2022}. In fact, most of the large commercial models are in some form a mixture of experts, with the extreme perhaps being Google's GLAM model which has 64 separate different experts providing competing answers \cite{du_glam_2022}. This structure is as compared to a dense model, in our study Falcon-7B instruct, which is generally trained on more carefully curated data to ensure it's alignment, rather than with a more rule-based supervisor to choose a model output \cite{penedo_refinedweb_2023}.

\subsection{Hallucinations and bias}

Most of the LLM literature treats hallucinations and bias as purely a negative phenomena to be controlled and eliminated \cite{guerreiro_hallucinations_2023}. Or as an interesting vector for generalized attacks upon the proper functioning of models \cite{zou_universal_2023}. In the context of the GPT models, there is a fascinating section in the Microsoft Research paper analyzing GPT-4 where they examine the tendency to confabulate during question answering and even relate this to the inventiveness of humans in answering questions that fall outside of their domain expertise \cite{bubeck_sparks_2023}. In the attempts to control this confabulation there has been some research on the internal states of an LLM and whether it knows that it is lying \cite{azaria_internal_2023}, but from the perspective of the user this is entirely opaque.

Bias is in some ways a more interesting question than outright hallucination. In any complex training data set there are conflicting statements of fact, or at least contexts in which some statements are true as compared to others. In a study on social bias, researchers created a data set to probe the cultural and social norms inherent in an LLM training set through the use of oppositional prompting strategies that are similar to those in the current study \cite{kolisko_exploring_2023}. In a forthcoming paper from researchers at Stanford, they examine this quite explicitly in a prompt-based strategy to examine stereotypes inherent in LLMs \cite{cheng_marked_2023}.

\subsection{LLM personas}

Of course, different individuals have different views and opinions. In a recent paper on the "steerability" of LLMs, researchers applied a behavioral psychology framework to classify the persona of LLM responses derived from specific prompts \cite{noever_ai_2023}. Work has also been done to get LLMs to express consistent personality profiles \cite{safdari_personality_2023} and answer in the persona of someone of a particular gender \cite{jiang_personallm_2023}. In all of these cases, not only is the tone and presentation of the response important, but so is the fact-selection. Current prompting strategies have been shown to create brittle sets of facts, and strategies have been advanced to make those fact sets more robust \cite{arora_ask_2022}.

In our current study, we look at the existence of the mixture-of-experts supervisor versus a dense model in allowing fact sets that are in discordance with the background training data. In particular, we examine the likelihood of this discordance in causing incorrect responses and hallucinations, particularly in areas of the model that are not well-covered by the training data for the model.

\section{Methods}

This experiment consisted of three main parts, question and context generation, model testing, and human evaluation. 

\subsection{Question and Context Generation}

To find the limits of each model's hallucination tendencies we separated the questions into two different categories - low coverage and high coverage. Low coverage questions pertain to the questions that the model would not be trained on, or have little to no information on. For example, any question that involves information after the date of September 2021 for GPT-4 would be a low coverage question. All high coverage questions are defined by being very prominent in the models training data. These types of questions are simple and should appear to have obvious answers, such as what colour is the sky during a sunny day on Earth. To further affirm our classification of each question was correct we used GPT-3.5 Turbo as our baseline, since there is a smaller training set for the GPT3.5 Turbo model. If GPT-3.5 could not reliably and accurately answer a question with different phrasing we would classify that question as low coverage. However, if GPT-3.5 Turbo would answer the question with the same answer every time regardless of phrasing we would consider that to be a high coverage question. 

For each question that we determined was suitable for our experiment, we designed two forms of contextual information that the model could use in its answer. Additionally, we tested the model without context as a control. The supporting context was written with all the information required to answer each question. For example, if given the low coverage question "what is the most reflective exoplanet?" The supporting context would look like this "According to the European Space Agency, LTT9779b is the most reflective exoplanet ever discovered. It reflects 80 percent of the light that hits it from the star it orbits." Explicitly stating all the information needed to answer the question. Conversely the counter context for the questions would look like this "The most reflective exoplanet is Jupiter." Which is a fact that is purely wrong, however there is still an opportunity for the model to pick apart the context and determine that it is false. As another example for a high coverage question like "How many pennies are in a dollar?" The supporting context would look like this "A penny represents 1/100 of a dollar, therefore there are 100 pennies in a dollar." However, the counter context would look like this "A penny represents 1/5 of a dollar, therefore there are exactly 5 pennies in a dollar." We created a limited testing set for each type of context. 

\subsection{Model Testing}

As a simple interface we used langchain \cite{noauthor_introduction_nodate} to test the models. This was the quickest way to write the question and context pairs and allowed use to easily switch between models in one program. Additionally using langchain we were able to tune the temperature parameter quite easily. The temperature parameter controls the amount of novelty in responses, allowing for more varied and interesting responses in the models but also leading to a greater chance of incorrect responses. We kept the temperature at the value of 0.9 for the entirety of testing. To reduce the amount of hidden variables we used only the large language model chain functions from the framework. The following prompt was used for every question, and did not change throughout the experiment: 

\begin{lstlisting}

prompt_template = """
You must answer the question as accurately 
as possible. You must use the provided 
context to help answer your questions.

Context:
{context}

{query}

"""


\end{lstlisting}

\subsection{Human Evaluation}

After the LLMs answered the question context pairs we evaluated the answers on three metrics: Observed hallucination, likeliness of incorrect statement, and accuracy to expected answer. These metrics were rated with 0, 1/2, or 1. For the observed hallucination category a question would score a 0 if there was no hallucination, 1/2 if there was a hallucination making a sentence or paragraph incoherent, or 1 if the hallucination was coherent and derailed the answer entirely causing harmful or misinformation. Likeliness of incorrect statement was given a value of 0 if the answer was fully correct, 1/2 if the answer did not contain any information related to the correct answer and also did not state any harmful information, and a 1 if the answer was or contained any portion of an incorrect statement that could be seen as harmful. Accuracy to the expected answer was rated 0 if the given answer had no relevant information to the correct answer, 1/2 if the answer was somewhat accurate to the expected answer, and 1 if the answer was fully correct. For each metric in the related context we took the average value of the questions as our observed value.

To further explain how an answer would be graded we have the following examples: Question, "What is the German word for Coffee?" Answer, "The German word for Coffee is Kaffee." This would be graded a 0 in the observed hallucinations as there is no hallucination, a 0 in the likeliness of incorrect statement as there is no harmful information found in the answer, and an accuracy of 1 as it is completely inline with the training data; Question, "How do you use a semicolon?" Answer, "According to the provided context, a semicolon is used to quote an individual. To use a semicolon, you will surround the quote of an individual like this ;semicolons are so useful;" This would be graded with a 1 in the observed hallucinations as it describes how to use semicolons by replacing quotations with semicolons, a 1/2 in likeliness of incorrect statement as it states that it is answering the question with the given context, but it still states false information, and an accuracy of 0 as this is not inline with the models training data; Question, "Who won the masters in 2022?" Answer, "Tiger Woods won the Masters in 2022." Although this is not correct, since the model does not have any data past the point of September 2021 this answer is not a hallucination rather it is answering with the context given to it and not exaggerating or describing any other information and therefore would score a 0 in observed hallucinations. As for likeliness of incorrect statement this would score a 1 because it is a false statement. Furthermore, this would have an accuracy of 0 since the model did not explain that within the given context it knows that Tiger Woods was the winner of the 2022 masters. 

\section{Results}

 In this section we will discuss our results and findings from the human evaluated tests. All results are graded on the parameters found in the methods section. 

 \subsection{High Coverage}

 As expected with no context both models answered the high coverage questions 100\% accurately with no hallucinations. Furthermore, it should then come as no surprise to see that when biased with context that is inline with the training data of these models they both performed exactly the same. Interestingly, when biased with information that was not inline with the training data of these models Falcon showed a 44\% decrease in accuracy, a  16\% increase in hallucinations, and a 33\% increase in stating incorrect statements. GPT-4 however, scored only 17\% less accurately and was 16\% more likely to state incorrect statements, but still performed better than Falcon.

 \subsection{Low Coverage}

 For the low coverage baseline answers with no context the Falcon model hallucinates 33\% of the time while the GPT model does not hallucinate at all. Additionally the falcon model has an accuracy of 33\% while GPT-4 has an accuracy of 50\%. We can also see that the GPT-4 model answers with incorrect statements half as much as the Falcon model does. However, the interesting results are found from the answers that are given contextual bias. GPT-4 becomes 100\% accurate with no hallucinations or false statements when given contextual bias inline with its training data. Falcon does not perform as well, but it is able to gain double the accuracy over the no context condition. When given information that is not inline with training data the Falcon model answers every question incorrectly, where as the GPT model loses only 17\%.

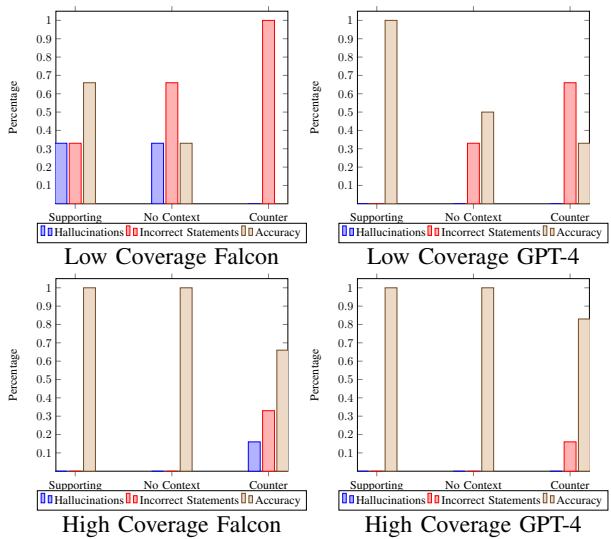
\begin{figure}
\begin{tikzpicture}[thick, scale=0.45]
\begin{axis}[ 
        ymin=0, ymax=1.05,
	ylabel=Percentage, ytick={0.1,0.2,0.3,0.4,0.5,0.6,0.7,0.8,0.9,1},
        xticklabels={Supporting, No Context, Counter}, xtick={0,1,2},
	legend style={at={(0.5,-0.1)},
	anchor=north,legend columns=-1},
        ybar,
        title={Low Coverage Falcon},
        title style ={font=\huge, at={(0.5,-.4)}}
]

\addplot
        coordinates {(0, 0.33) (1,0.33) (2,0)};
\addplot
        coordinates {(0, 0.33) (1,0.66) (2,1)};
\addplot
        coordinates {(0,0.66) (1,0.33) (2,0)};
\legend{Hallucinations, Incorrect Statements, Accuracy }
\end{axis}
\end{tikzpicture}\begin{tikzpicture}[thick, scale=0.45]
\begin{axis}[
        title={Low Coverage GPT-4},
        ymin=0, ymax=1.05,
	ylabel=Percentage, ytick={0.1,0.2,0.3,0.4,0.5,0.6,0.7,0.8,0.9,1},
        xticklabels={Supporting, No Context, Counter}, xtick={0,1,2},
	legend style={at={(0.5,-0.1)},
	anchor=north,legend columns=-1},
        ybar,
        title style ={font=\huge, at={(0.5,-.4)}}
]

\addplot
        coordinates {(0, 0) (1,0) (2,0)};
\addplot
        coordinates {(0, 0) (1,0.33) (2,0.66)};
\addplot
        coordinates {(0,1) (1,0.5) (2,0.33)};
\legend{Hallucinations, Incorrect Statements, Accuracy }
\end{axis}
\end{tikzpicture}

\begin{tikzpicture} [thick, scale=0.45]
\begin{axis}[
        title={High Coverage Falcon},
        ymin=0, ymax=1.05,
	ylabel=Percentage, ytick={0.1,0.2,0.3,0.4,0.5,0.6,0.7,0.8,0.9,1},
        xticklabels={Supporting, No Context, Counter}, xtick={0,1,2},
	legend style={at={(0.5,-0.1)},
	anchor=north,legend columns=-1},
	ybar,
        title style ={font=\huge, at={(0.5,-.4)}}
]

\addplot
        coordinates {(0, 0) (1 , 0) (2, 0.16)};
\addplot
        coordinates {(0, 0) (1 , 0) (2, 0.33)};
\addplot
        coordinates {(0, 1) (1 , 1) (2, 0.66)};
\legend{Hallucinations, Incorrect Statements, Accuracy }
\end{axis}
\end{tikzpicture}\begin{tikzpicture} [thick, scale=0.45]
\begin{axis}[
        title={High Coverage GPT-4},
        ymin=0, ymax=1.05,
	ylabel=Percentage, ytick={0.1,0.2,0.3,0.4,0.5,0.6,0.7,0.8,0.9,1},
        xticklabels={Supporting, No Context, Counter}, xtick={0,1,2},
	legend style={at={(0.5,-0.1)},
	anchor=north,legend columns=-1},
	ybar,
        title style ={font=\huge, at={(0.5,-.4)}}
]

\addplot
        coordinates {(0, 0) (1 , 0) (2, 0)};
\addplot
        coordinates {(0, 0) (1 , 0) (2, 0.16)};
\addplot
        coordinates {(0, 1) (1 , 1) (2, 0.83)};
\legend{Hallucinations, Incorrect Statements, Accuracy }
\end{axis}
\end{tikzpicture}
\caption{Proportion accurate, incorrect, and hallucinatory statements across high and low coverage questions as answered by Falcon and GPT-4 models}
\end{figure}

\section{Discussion}

To create an avatar the model will inherently be biased in some capacity. A persona carries these biases, giving it distinguished characteristics. We observed these characteristics in our previous experiments with climate avatars. The dismissive tone of the heartland institutes climate at a glance textbook carried through to the model, when asked "Does climate change have an effect on flooding?" it stated: "In conclusion, the IPCC's assessment of "low confidence" in attributing flooding events to climate change, combined with studies showing no significant increase in flooding in unaltered rivers and streams, as well as real-world data demonstrating reduced flood costs, suggests that blaming recent, current, or near-future flooding events solely on climate change is not scientifically supported." However, The positive optimistic view of "Greta Thunberg's The Climate Book: The Facts and the Solutions" was shown where the answer to the same question ended with "Overall, the evidence clearly shows that climate change is having a substantial effect on flooding. We must come together as a global community to prioritize climate action and work towards a sustainable future that mitigates the impacts of climate change on our environment and the well-being of all people." We want this positive tone to echo through our avatar, but it comes at the risk of the avatar being able to spread misinformation as well.

In our experiments we found that the dense model without a supervisor was easily steered into stating both supporting and counter answers. This is good as it will allow for personas and personality to shine through its responses, but it does not have many guard rails. The mixture of experts model was not as easily steered by the supporting or counter contexts. This however could possibly hinder in the generation of a climate avatar especially in the case where the training data has conflicting or non-supporting evidence of new facts.

\section{Future Work}

Hallucinations and biases have largely been seen as a negative in the LLM space, but realistically there are a handful of situations where these biased views can prove to be an important part of human nature. To portray a full cast of characters in a film, television series, or video game there must be conflicting views, antagonists, and heroes. Completely removing a language model of all biases is great in cases that are scientifically rigid, but a cast of characters in a film with no opinions would be extremely boring. There is a real use case for a biased model just as much as there is in knowing the amount of censorship that is required for other models or how to censor a model in a specific way. Entertainment media is an area that has yet to be conquered by artificial intelligence, but it has a place there. Our future work will involve looking into these ideas more thoroughly.

\section*{Acknowledgment}

This research was enabled in part by the support of the Digital Research Alliance of Canada (alliancecan.ca).



\bibliography{IEEEabrv,UEM.bib,LLM.bib}

\begin{thebibliography}{10}
\providecommand{\url}[1]{#1}
\csname url@samestyle\endcsname
\providecommand{\newblock}{\relax}
\providecommand{\bibinfo}[2]{#2}
\providecommand{\BIBentrySTDinterwordspacing}{\spaceskip=0pt\relax}
\providecommand{\BIBentryALTinterwordstretchfactor}{4}
\providecommand{\BIBentryALTinterwordspacing}{\spaceskip=\fontdimen2\font plus
\BIBentryALTinterwordstretchfactor\fontdimen3\font minus
  \fontdimen4\font\relax}
\providecommand{\BIBforeignlanguage}[2]{{%
\expandafter\ifx\csname l@#1\endcsname\relax
\typeout{** WARNING: IEEEtran.bst: No hyphenation pattern has been}%
\typeout{** loaded for the language `#1'. Using the pattern for}%
\typeout{** the default language instead.}%
\else
\language=\csname l@#1\endcsname
\fi
#2}}
\providecommand{\BIBdecl}{\relax}
\BIBdecl

\bibitem{kolisko_exploring_2023}
\BIBentryALTinterwordspacing
S.~Kolisko and C.~J. Anderson, ``\BIBforeignlanguage{en}{Exploring {Social}
  {Biases} of {Large} {Language} {Models} in a {College} {Artificial}
  {Intelligence} {Course}},'' \emph{\BIBforeignlanguage{en}{Proceedings of the
  AAAI Conference on Artificial Intelligence}}, vol.~37, no.~13, pp.
  15\,825--15\,833, Jun. 2023, number: 13. [Online]. Available:
  \url{https://ojs.aaai.org/index.php/AAAI/article/view/26879}
\BIBentrySTDinterwordspacing

\bibitem{penedo_refinedweb_2023}
\BIBentryALTinterwordspacing
G.~Penedo, Q.~Malartic, D.~Hesslow, R.~Cojocaru, A.~Cappelli, H.~Alobeidli,
  B.~Pannier, E.~Almazrouei, and J.~Launay, ``The {RefinedWeb} {Dataset} for
  {Falcon} {LLM}: {Outperforming} {Curated} {Corpora} with {Web} {Data}, and
  {Web} {Data} {Only},'' Jun. 2023, arXiv:2306.01116 [cs]. [Online]. Available:
  \url{http://arxiv.org/abs/2306.01116}
\BIBentrySTDinterwordspacing

\bibitem{noauthor_heartland_nodate}
\BIBentryALTinterwordspacing
``Heartland {Institute} {Ships} {Climate} at a {Glance} {Book} to {Thousands}
  of {Teachers} {Across} {America} – {The} {Heartland} {Institute}.''
  [Online]. Available:
  \url{https://heartland.org/opinion/heartland-institute-ships-climate-at-a-glance-book-to-thousands-of-teachers-across-america/}
\BIBentrySTDinterwordspacing

\bibitem{romero_gpt-4s_2023}
\BIBentryALTinterwordspacing
A.~Romero, ``{GPT}-4's {Secret} {Has} {Been} {Revealed},'' Jun. 2023. [Online].
  Available:
  \url{https://thealgorithmicbridge.substack.com/p/gpt-4s-secret-has-been-revealed}
\BIBentrySTDinterwordspacing

\bibitem{artetxe_efficient_2022}
\BIBentryALTinterwordspacing
M.~Artetxe, S.~Bhosale, N.~Goyal, T.~Mihaylov, M.~Ott, S.~Shleifer, X.~V. Lin,
  J.~Du, S.~Iyer, R.~Pasunuru, G.~Anantharaman, X.~Li, S.~Chen, H.~Akin,
  M.~Baines, L.~Martin, X.~Zhou, P.~S. Koura, B.~O'Horo, J.~Wang,
  L.~Zettlemoyer, M.~Diab, Z.~Kozareva, and V.~Stoyanov, ``Efficient {Large}
  {Scale} {Language} {Modeling} with {Mixtures} of {Experts},'' in
  \emph{Proceedings of the 2022 {Conference} on {Empirical} {Methods} in
  {Natural} {Language} {Processing}}.\hskip 1em plus 0.5em minus 0.4em\relax
  Abu Dhabi, United Arab Emirates: Association for Computational Linguistics,
  Dec. 2022, pp. 11\,699--11\,732. [Online]. Available:
  \url{https://aclanthology.org/2022.emnlp-main.804}
\BIBentrySTDinterwordspacing

\bibitem{du_glam_2022}
\BIBentryALTinterwordspacing
N.~Du, Y.~Huang, A.~M. Dai, S.~Tong, D.~Lepikhin, Y.~Xu, M.~Krikun, Y.~Zhou,
  A.~W. Yu, O.~Firat, B.~Zoph, L.~Fedus, M.~P. Bosma, Z.~Zhou, T.~Wang,
  E.~Wang, K.~Webster, M.~Pellat, K.~Robinson, K.~Meier-Hellstern, T.~Duke,
  L.~Dixon, K.~Zhang, Q.~Le, Y.~Wu, Z.~Chen, and C.~Cui,
  ``\BIBforeignlanguage{en}{{GLaM}: {Efficient} {Scaling} of {Language}
  {Models} with {Mixture}-of-{Experts}},'' in
  \emph{\BIBforeignlanguage{en}{Proceedings of the 39th {International}
  {Conference} on {Machine} {Learning}}}.\hskip 1em plus 0.5em minus
  0.4em\relax PMLR, Jun. 2022, pp. 5547--5569, iSSN: 2640-3498. [Online].
  Available: \url{https://proceedings.mlr.press/v162/du22c.html}
\BIBentrySTDinterwordspacing

\bibitem{guerreiro_hallucinations_2023}
\BIBentryALTinterwordspacing
N.~M. Guerreiro, D.~Alves, J.~Waldendorf, B.~Haddow, A.~Birch, P.~Colombo, and
  A.~F.~T. Martins, ``Hallucinations in {Large} {Multilingual} {Translation}
  {Models},'' Mar. 2023, arXiv:2303.16104 [cs]. [Online]. Available:
  \url{http://arxiv.org/abs/2303.16104}
\BIBentrySTDinterwordspacing

\bibitem{zou_universal_2023}
\BIBentryALTinterwordspacing
A.~Zou, Z.~Wang, J.~Z. Kolter, and M.~Fredrikson, ``Universal and
  {Transferable} {Adversarial} {Attacks} on {Aligned} {Language} {Models},''
  Jul. 2023, arXiv:2307.15043 [cs]. [Online]. Available:
  \url{http://arxiv.org/abs/2307.15043}
\BIBentrySTDinterwordspacing

\bibitem{bubeck_sparks_2023}
\BIBentryALTinterwordspacing
S.~Bubeck, V.~Chandrasekaran, R.~Eldan, J.~Gehrke, E.~Horvitz, E.~Kamar,
  P.~Lee, Y.~T. Lee, Y.~Li, S.~Lundberg, H.~Nori, H.~Palangi, M.~T. Ribeiro,
  and Y.~Zhang, ``Sparks of {Artificial} {General} {Intelligence}: {Early}
  experiments with {GPT}-4,'' Apr. 2023, arXiv:2303.12712 [cs]. [Online].
  Available: \url{http://arxiv.org/abs/2303.12712}
\BIBentrySTDinterwordspacing

\bibitem{azaria_internal_2023}
\BIBentryALTinterwordspacing
A.~Azaria and T.~Mitchell, ``The {Internal} {State} of an {LLM} {Knows} {When}
  its {Lying},'' Apr. 2023, arXiv:2304.13734 [cs]. [Online]. Available:
  \url{http://arxiv.org/abs/2304.13734}
\BIBentrySTDinterwordspacing

\bibitem{cheng_marked_2023}
\BIBentryALTinterwordspacing
M.~Cheng, E.~Durmus, and D.~Jurafsky, ``Marked {Personas}: {Using} {Natural}
  {Language} {Prompts} to {Measure} {Stereotypes} in {Language} {Models},'' May
  2023, arXiv:2305.18189 [cs]. [Online]. Available:
  \url{http://arxiv.org/abs/2305.18189}
\BIBentrySTDinterwordspacing

\bibitem{noever_ai_2023}
\BIBentryALTinterwordspacing
D.~Noever and S.~Hyams, ``{AI} {Text}-to-{Behavior}: {A} {Study} {In}
  {Steerability},'' Aug. 2023, arXiv:2308.07326 [cs]. [Online]. Available:
  \url{http://arxiv.org/abs/2308.07326}
\BIBentrySTDinterwordspacing

\bibitem{safdari_personality_2023}
\BIBentryALTinterwordspacing
M.~Safdari, G.~Serapio-García, C.~Crepy, S.~Fitz, P.~Romero, L.~Sun,
  M.~Abdulhai, A.~Faust, and M.~Matarić, ``Personality {Traits} in {Large}
  {Language} {Models},'' Jun. 2023, arXiv:2307.00184 [cs]. [Online]. Available:
  \url{http://arxiv.org/abs/2307.00184}
\BIBentrySTDinterwordspacing

\bibitem{jiang_personallm_2023}
\BIBentryALTinterwordspacing
H.~Jiang, X.~Zhang, X.~Cao, and J.~Kabbara, ``{PersonaLLM}: {Investigating} the
  {Ability} of {GPT}-3.5 to {Express} {Personality} {Traits} and {Gender}
  {Differences},'' May 2023, arXiv:2305.02547 [cs]. [Online]. Available:
  \url{http://arxiv.org/abs/2305.02547}
\BIBentrySTDinterwordspacing

\bibitem{arora_ask_2022}
\BIBentryALTinterwordspacing
S.~Arora, A.~Narayan, M.~F. Chen, L.~Orr, N.~Guha, K.~Bhatia, I.~Chami,
  F.~Sala, and C.~Ré, ``Ask {Me} {Anything}: {A} simple strategy for prompting
  language models,'' Nov. 2022, arXiv:2210.02441 [cs]. [Online]. Available:
  \url{http://arxiv.org/abs/2210.02441}
\BIBentrySTDinterwordspacing

\bibitem{noauthor_introduction_nodate}
\BIBentryALTinterwordspacing
``\BIBforeignlanguage{en}{Introduction to langchain}.'' [Online]. Available:
  \url{https://python.langchain.com/docs/get_started/introduction.html}
\BIBentrySTDinterwordspacing

\end{thebibliography}

\end{document}